\RequirePackage[svgnames,table]{xcolor}

\documentclass[11pt,letterpaper,logo]{yalearxiv}

\usepackage{graphicx}
\usepackage{float,epstopdf}
\usepackage{bbm}

\usepackage{microtype}

\usepackage{natbib}
\setcitestyle{square}

\usepackage{subcaption}
\usepackage{booktabs}
\usepackage{amsmath}
\usepackage{amssymb}
\usepackage{mathtools}
\usepackage{amsthm}
\usepackage{dsfont}
\usepackage{multicol}
\usepackage{makecell}
\usepackage{multirow}
\usepackage{amsfonts}
\usepackage{mathrsfs}
\usepackage[amssymb, thickqspace]{SIunits}
\usepackage{enumitem}
\usepackage{pgfplotstable}
\usepackage{lipsum}		

\usepackage{microtype}
\usepackage{graphicx}
\usepackage{booktabs} 
\usepackage[table]{xcolor}
\usepackage{arydshln}
\usepackage[normalem]{ulem} 

\usepackage{cases}
\usepackage{wrapfig}

\usepackage{url}

\usepackage{thmtools}
\usepackage{thm-restate}
\usepackage{tabu}

\definecolor{huskypurple}{HTML}{4B2E83}

\usepackage{titletoc}
\usepackage{listings}
\lstdefinestyle{promptstyle}{
  basicstyle=\ttfamily\footnotesize,
  breaklines=true,
  breakautoindent=false,
  breakindent=0pt,
  postbreak=\mbox{\textcolor{gray}{$\hookrightarrow$}\space},
  columns=fullflexible,
  keepspaces=true,
  frame=single,
  framesep=5pt,
  xleftmargin=6pt,
  xrightmargin=6pt,
  aboveskip=8pt,
  belowskip=8pt,
  showstringspaces=false,
}

\usepackage{array}
\usepackage{tikz}
\usetikzlibrary{arrows.meta,positioning,decorations.pathreplacing,calc,shapes.geometric,fit,backgrounds,shadows}
\usepackage{algorithm}
\usepackage{algpseudocode}

\makeatletter
\def\munderbar#1{\underline{\sbox\tw@{$#1$}\dp\tw@\z@\box\tw@}}
\makeatother

\newcommand{\be}{\begin{equation}}
\newcommand{\ee}{\end{equation}}
\newcommand{\bea}{\begin{equation*}\begin{aligned}}
\newcommand{\eea}{\end{aligned}\end{equation*}}

\newcommand{\mbb}{\mathbb}

\newcommand{\p}{\mbb P}

\newcommand{\x}{x}

\newcommand{\Mpool}{\mathcal{M}}
\newcommand{\Hist}{\mathcal{H}}
\newcommand{\Acts}{\mathcal{A}}
\newcommand{\Agui}{\mathcal{A}_{\mathrm{gui}}}
\newcommand{\Alook}{\mathcal{A}_{\mathrm{look}}}
\newcommand{\Acode}{\mathcal{A}_{\mathrm{code}}}
\newcommand{\Ain}{\mathcal{A}_{\mathrm{in}}}
\newcommand{\Aro}{\mathcal{A}_{\mathrm{ro}}}
\newcommand{\Bgui}{B_{\mathrm{gui}}}
\newcommand{\Nsave}{N_{\mathrm{save}}}
\newcommand{\Nlook}{N_{\mathrm{look}}}
\newcommand{\Nadv}{N_{\mathrm{adv}}}
\newcommand{\Nblock}{N_{\mathrm{block}}}
\newcommand{\Lib}{\mathcal{G}}
\definecolor{ctxc}{HTML}{3A6EA5}   
\definecolor{prbc}{HTML}{3C8D5A}   
\definecolor{verc}{HTML}{C0762B}   
\definecolor{grayc}{HTML}{8A8F98}
\definecolor{failc}{HTML}{B5443B}

\newcommand{\figfn}{\fontsize{8}{9.6}\selectfont}
\newcommand{\figsc}{\fontsize{7}{8.4}\selectfont}
\newcommand{\figti}{\fontsize{5}{6}\selectfont}

\tikzset{
  fig/.style={font=\sffamily\figfn, >={Stealth[length=4pt]}},
  panel/.style={rounded corners=6pt, fill=#1!6, draw=#1!35, line width=0.6pt},
  phead/.style={font=\sffamily\figfn\bfseries, text=#1!85!black, anchor=north west},
  card/.style={rounded corners=2.5pt, fill=white, draw=black!22, line width=0.5pt, align=flush left, font=\sffamily\figsc,
               inner sep=4pt, drop shadow={opacity=0.12, shadow xshift=0.4pt, shadow yshift=-0.6pt}},
  code/.style={font=\ttfamily\figsc},
  tag/.style={circle, fill=black!9, draw=black!25, inner sep=0.6pt, minimum size=0.42cm, font=\figfn},
  lbl/.style={font=\sffamily\figsc, text=black!60, align=flush center},
  arr/.style={->, line width=0.8pt, black!55},
  badge/.style={circle, fill=#1, text=white, inner sep=0pt, minimum size=0.34cm, font=\sffamily\figsc\bfseries},
  banner/.style={rounded corners=3pt, fill=black!5, draw=black!18, inner sep=4pt, font=\sffamily\figfn},
  win/.pic={
    \draw[rounded corners=0.6pt, fill=white, draw=black!45, line width=0.35pt] (-0.3,-0.21) rectangle (0.3,0.21);
    \fill[black!30] (-0.3,0.16) rectangle (0.3,0.21);
    \fill[black!12] (-0.26,-0.17) rectangle (-0.14,0.12);
    \fill[ctxc!25] (-0.1,-0.17) rectangle (0.26,0.12);
    \draw[black!50, line width=0.3pt] (-0.04,-0.1) .. controls (0.04,0.02) and (0.1,-0.08) .. (0.2,0.0);},
  wingone/.pic={
    \draw[rounded corners=0.6pt, densely dashed, draw=black!35, fill=black!3] (-0.3,-0.21) rectangle (0.3,0.21);
    \node[text=black!45, font=\figsc] at (0,0) {$\rho$};},
  txtreply/.pic={
    \draw[rounded corners=0.6pt, fill=white, draw=black!45, line width=0.35pt] (-0.3,-0.21) rectangle (0.3,0.21);
    \foreach \y in {0.1,0.02,-0.06,-0.14} \draw[black!35, line width=0.4pt] (-0.22,\y) -- (0.18,\y);
    \node[font=\ttfamily\figti, text=black!60] at (-0.21,0.1) {};},
}

\newtcolorbox{simpleElegantQuote}{
    colback=AliceBlue!50!White,
    colframe=RoyalBlue!75!Black,
    boxrule=0.5pt,
    arc=2mm,
    boxsep=4pt,
    left=10pt, right=10pt,
    top=8pt, bottom=8pt,
    fontupper=\itshape,
}

\title{Towards Hyper Agents for Long-Horizon Computer Use: Harness Mechanisms in openJiuwen}

\runningtitle{Towards Hyper Agents for Long-Horizon Computer Use}


\usepackage{fontawesome5}   
\usepackage{float}
\usepackage{xurl}   
\usepackage{needspace}

\definecolor{yaleblue}{RGB}{0,58,112}

\author{openJiuwen Team\\
Huawei Technologies Co., Ltd.\\
\faGithub~\textbf{Source Code:} \href{https://github.com/openJiuwen-ai/computer-use/tree/osworld-challenge}{\texttt{github.com/openJiuwen-ai/computer-use}}}

\hypersetup{colorlinks=true, linkcolor=blue!50!black, citecolor=blue!50!black,
            urlcolor=blue!50!black}

\begin{document}

\begin{abstract}
\vspace{-1mm}
\begin{center}
    \section*{Abstract}
\end{center}

Computer-use agents operate real desktop applications through the screen, and a single task can take dozens of steps across several applications. Over such long horizons, three challenges recur: \textbf{state and context drift}, as screenshots crowd out the instruction and the progress that matters most never appears on screen; a \textbf{knowledge gap}, as the agent starts from a wrong picture of the installed software, the named resources and whether the task can be done at all; and \textbf{error accumulation}, as GUI outcomes are only partly observable and an uncaught error becomes the starting state of every later step. We propose a set of harness mechanisms that address these challenges around one fixed general model, with no training and no extra model calls in the main loop. \textbf{State and Context Management} keeps task-wide facts in fixed positions, bounds the screenshots in view and tracks for the model the budget, unsaved work and repeated failures it cannot see. \textbf{Probing Before Acting} checks the task against the machine in a read-only pass before the first action and ends a task as infeasible only with evidence. \textbf{Verifying After Acting} makes every desktop action report what it actually did, with an explicit mark and a fresh capture of its window. Implemented on openJiuwen's agent-core and evaluated on \textbf{OSWorld-Verified}, the harness achieves \textbf{93.5\%}, \textbf{3.3 percentage points} above the strongest leaderboard entry and \textbf{10.1 percentage points} above Anthropic's own agent for the same model, and removing any one mechanism lowers accuracy or raises cost.

\end{abstract}

\maketitle

\section{Introduction}\label{sec:intro}

Large multimodal models are turning computer use from scripted automation into general-purpose GUI agents: agents that look at the screen, then click, type and scroll through real applications to complete open-ended tasks. Most software is built for people rather than programs, and much of it has no API, so an agent that can operate the interface directly reaches work no integration can. Doing so reliably is a long-horizon problem. A single desktop task can take dozens of steps across several applications, and the agent must stay coherent while the screen, the files on disk and its own progress change throughout the run. On OSWorld, a benchmark of such tasks on real desktops, the best model at its release completed 12.2\% of tasks against 72.4\% for people~\citep{xie2024osworld}.

Recent GUI agents close much of this gap through distinct architectural choices: native models trained end to end, compositional frameworks that split planning from grounding, and multi-agent systems that mix GUI actions with code execution. Together, these systems show that a GUI agent is no longer a model with a screenshot tool: the system around the model increasingly decides how perception, knowledge and verification are assembled, and how each step of execution is controlled.

As this surrounding system grows, it becomes a systems problem of its own. We identify three challenges that a computer-use agent meets on any long task. They concern what the agent keeps in view while it works, what it knows before it acts, and what it learns after each action.

The first is \textbf{state and context drift}. Every step adds a screenshot, so the context grows with the task and stale images crowd out the instruction, the rules and the record of what has been done. Some of what the agent most needs is never on screen at all: how many steps it has left, whether its work has been saved, and whether it has already tried the same failing action. A model left to track this itself loses count, and keeps looking without acting, runs past any reasonable number of steps, or finishes work it never saves.

The second is the \textbf{knowledge gap}. A capable model often fails a task not for lack of skill but because it starts from a wrong picture of the task: which features the installed application has, whether the files, settings and devices the task names exist, and whether the task can be done at all. Knowledge offered on demand goes unread when the agent does not know what it lacks, so these beliefs are rarely checked before the first action, and an agent with a wrong belief attempts the impossible or gives up on the possible.

The third is \textbf{error accumulation}. The outcome of a GUI action is only partly observable: a click is read off an approximate view of the screen, input can reach a window other than the one intended, and a saved file can differ from what the application displays. An error that is not caught at the step where it happens becomes the starting state of later steps, so over long tasks, small errors compound into failure.

To address these challenges, we propose a new set of harness mechanisms that
tackle the main failure modes of computer-use agents (CUAs) on long-horizon
tasks, and implement them in a general computer-use agent built on openJiuwen's
agent-core~\citep{openjiuwen2026}. Rather than
training or specializing the model, we keep one general model fixed and place
the solution in the harness around it, a step towards \emph{hyper agents} whose
long-horizon reliability comes from the harness, not from the model alone. The
harness has three mechanisms, one for each challenge.
\textbf{State and Context Management} keeps state and context from drifting
over long trajectories: it bounds what the model sees and tracks for it what it
cannot see. \textbf{Probing Before Acting} closes the knowledge gap by
gathering and checking the task's knowledge in a read-only pass before the
first action. \textbf{Verifying After Acting} stops errors from accumulating
across steps by making every desktop action report what it actually did. We
evaluate the agent on OSWorld-Verified. Our contributions are:

\begin{itemize}
  \item \textbf{State and context management for long horizons.} We split the
  context into a persistent part that holds task-wide facts and a bounded
  window of recent perception, and recite the task briefing at the end of the
  context before every model call without adding it to the history, so the
  context stays bounded however long the task runs. Alongside the context, the
  harness keeps a state the model cannot observe (actions used, unsaved edits
  per window and repeated failures), reports the budget and unsaved work as
  transient notices, and warns about and then refuses a call that keeps
  failing the same way.
  \item \textbf{Evidence-gated probing before acting.} A read-only probing pass
  loads the application knowledge the task needs, probes the files and
  applications it names and decides feasibility, and may end a task as
  infeasible only with evidence. Every rule it quotes is expanded back to the
  paragraph it came from, so the rule keeps its condition.
  \item \textbf{Verified actions without extra model calls.} An input action
  is aimed in the pixels of a native-resolution capture of its window and is
  sent only if that window can be brought to the front. Its reply carries an
  explicit success or failure mark and, on success, a fresh capture of the
  window, so an error surfaces at the step where it happens instead of
  compounding over later steps, and the agent can read a saved file back from
  disk instead of judging it from the screen.
\end{itemize}

\section{Related Work}\label{sec:related_work}

\subsection{GUI-Agent Architectures}\label{sec:rw-arch}

Recent work has advanced GUI agents through different architectural choices. Commercial computer-use models, such as Claude computer use~\citep{anthropic2024computeruse} and OpenAI's Computer-Using Agent~\citep{openai2025cua}, operate interfaces from screenshots with a single general model, and reported 14.9\% and 38.1\% on OSWorld at release. Native agents train one model end to end: UI-TARS~\citep{qin2025uitars} learns from interaction traces collected across hundreds of virtual machines. A related line strengthens visual grounding, the step of locating the right on-screen element: OS-Atlas~\citep{wu2024osatlas} and UGround~\citep{gou2024uground} train grounding models on millions of interface elements, and GTA1~\citep{yang2025gta1} adds reinforcement-learned grounding and test-time selection among candidate actions. Compositional frameworks divide the work among models: Agent~S~\citep{agashe2024agents} and Agent~S2~\citep{agashe2025agents2} pair a planner with specialist grounding models, and CoAct-1~\citep{song2025coact} lets an orchestrator hand subtasks either to a GUI operator or to a programmer agent that writes and runs scripts, reaching 60.76\% on OSWorld-Verified. These approaches primarily improve the model, the grounding or the division of labour between agents. Comparatively less attention has been paid to the harness around a single general model as the layer that decides what the model sees, what it knows and what is checked at each step.

\subsection{Knowledge and Reflection at Run Time}

Several GUI agents supplement the model with knowledge gathered outside the current step. Agent~S retrieves external web knowledge and past experience to guide its plans, and Agent Workflow Memory~\citep{wang2024awm} induces reusable routines from past tasks and supplies them to later ones. Reflection gives agents a way to notice and correct their mistakes: Reflexion~\citep{shinn2023reflexion} stores verbal self-critique in memory instead of updating weights, and Mobile-Agent-v2~\citep{wang2024mobileagentv2} adds a dedicated reflection agent that checks the outcome of each operation. These works show the value of knowledge and checking beyond the model's own reasoning, but mostly add them as further model calls. Desktop tasks also offer cheaper signals that a harness can gather directly: the state of the machine before the task starts, read-only probes of the files and applications a task names, and the saved result itself. Feasibility in particular is rarely a step of its own: most GUI agents judge it in passing, through prompt rules or an output label (CoAct-1, GTA1), and ConflictGuard~\citep{huang2026conflictguard} adds a feasibility check before acting. Our agent makes it a separate read-only pass that may end a task only with evidence.

\subsection{Context Management for Long-Horizon Agents}

Another line of work manages long-horizon execution by controlling what remains visible to the model. For coding agents, simply masking older observations roughly halves cost while matching LLM-based summarisation~\citep{lindenbauer2025complexity}. GUI agents face a sharper version of the problem, because each observation is a screenshot. Mobile-Agent-v2 keeps a memory unit of task-relevant content as progress changes, and hierarchical planners such as Agent~S2 keep the overall plan separate from step-level detail. Where information sits in the context matters as well as whether it is kept: models attend less to the middle of long inputs~\citep{liu2024lost}, and long-running agents tend to drift from their original goal~\citep{arike2025goaldrift}. Manus counters both by restating the objective at the end of the context, a practice it calls recitation~\citep{ji2025manus}. These methods adapt what the model is shown, but long GUI tasks also need the harness to track progress the model loses track of: how many steps remain, whether work has been saved, and whether the agent is repeating itself.

\begin{figure}[t]
  \centering
  \includegraphics[width=\linewidth]{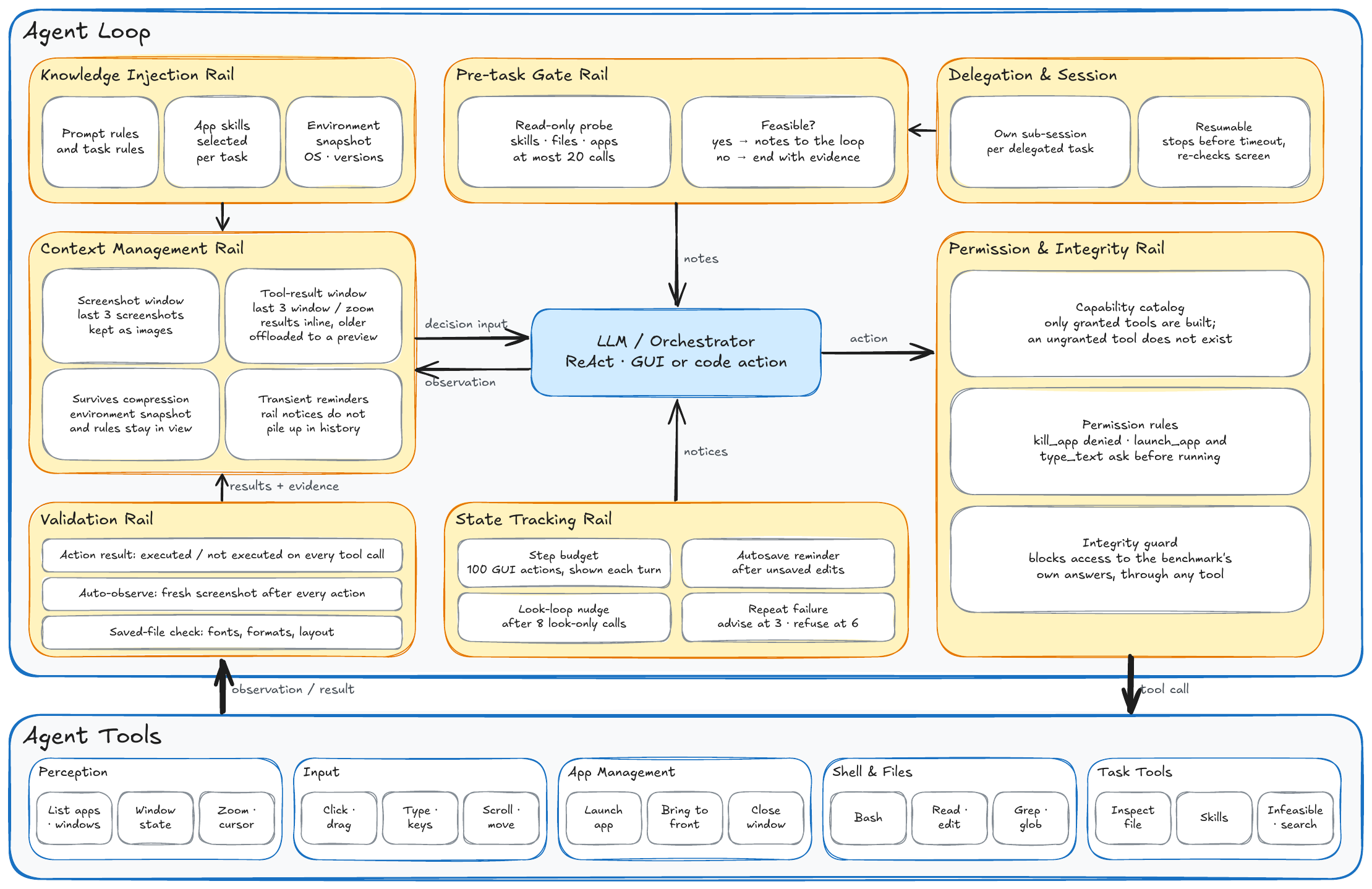}
  \caption{\textbf{Architecture of our agent.} Inside the agent loop, rails around the LLM manage knowledge, context, state, feasibility and validation. Every tool call passes the rails before it reaches the tools.}
  \label{fig:arch}
\end{figure}

\paragraph{Positioning of our work.} Our agent differs from the work above in four ways.

\begin{itemize}
  \item \textbf{One model, GUI and code in one loop.} CoAct-1 and Agent~S2 split the work between an orchestrator or planner and separate GUI and coding agents. Our agent chooses between a click and a shell command at every step of one loop, so no task is split between agents in advance.
  \item \textbf{The harness tracks state the model cannot see.} Masking, memory units and hierarchical plans change what the model is shown. Our harness also counts what no observation shows: GUI actions left in the budget, unsaved edits in each window, and identical failed calls. It tells the model when one of these matters, and refuses a call that keeps failing the same way.
  \item \textbf{Task knowledge is gathered before acting, not on demand.} Agent~S and Agent Workflow Memory retrieve web knowledge or past routines while the agent works. Our agent runs a read-only pass before the first action that loads the application skills the task needs, probes the files and applications it names, and decides feasibility, ending the task only when it has evidence. Each quoted rule is expanded to its full paragraph, so it keeps its condition.
  \item \textbf{Checking costs no extra model calls.} Reflexion and Mobile-Agent-v2 check outcomes with further model calls, such as self-critique or a reflection agent. Our agent checks with its tools: every reply carries a success or failure mark, input is sent only to a window that can be brought to the front and returns a fresh capture of it, and saved files are read back from disk.
\end{itemize}

\section{Overall Architecture}\label{sec:architecture}

Our agent is a single reasoning loop that looks at the screen, decides on one action, carries it out and looks again, until the task is done. It lives in openJiuwen's agent-core. Figure~\ref{fig:arch} shows the system.

\paragraph{Loop and tools.} The agent has two parts. The \emph{agent loop} holds the model and the rails, openJiuwen's mechanism for attaching behaviour at fixed points in the loop: before each model call, before each tool call and after each tool call. The \emph{tools} give the model perception, input, application management, shell and file access, and task-level tools such as a structured file inspector and application skills. Every tool call passes the rails before it reaches the tools.

\paragraph{Pipeline.} Each task passes through three stages:

\begin{enumerate}
  \item \textbf{Snapshot.} Before the agent starts, it records the operating system, the installed application versions and editor extensions, the open windows, the files on the desktop and in the home directory, and the current date.
  \item \textbf{Gate.} A read-only agent reads the task, loads the relevant application knowledge and checks the resources the task names. If it finds clear evidence that the task cannot be done, it ends the task as infeasible. Otherwise it passes its notes on.
  \item \textbf{Main agent.} The main agent works through the task with GUI actions (clicking, typing, scrolling) and code actions (shell commands, reading and editing files), choosing whichever suits the step. Every action it takes passes the rails that watch and steer it. It stops when it answers without calling a tool or when its GUI action budget runs out, and the result is the state it left on the machine. A task it judges impossible is reported with a dedicated tool, after which it stops in the same way.
\end{enumerate}

\paragraph{Design highlights.}

\begin{itemize}
  \item \textbf{Every action is visible.} The model can act only through tools that pass the rails, so nothing it does escapes monitoring or step budgets, and a call can be refused before it reaches the machine.
  \item \textbf{Code and GUI as equals.} The agent uses code when the result lives in a file and the GUI when an open application must hold the result.
  \item \textbf{Check the result, not the screen.} Saved files are read back with a structured inspector rather than judged from screenshots.
\end{itemize}

\paragraph{System prompt.} The system prompt is assembled from the base prompt of agent-core's computer-use agent, guidance on choosing between the shell, the file tools and the GUI, a note on operations that need administrator rights, about two dozen cross-application working rules, and a short set of rules, shared with the gate, on judging feasibility and completion and on how to carry out a feasible task. Most rules were written in response to an observed failure and are stated in general terms.

\section{Methodology}\label{sec:method}

\subsection{Problem Formulation}\label{sec:formulation}

A task gives the agent an instruction $q$ and a machine whose state $s_t$ it can observe and change only through its tools, so the agent acts under partial observability. At each step $t$ the harness builds a context $c_t$ from the information available to it, $\Mpool_t$, and a fixed model $\pi$ chooses an action:
\begin{equation}\label{eq:loop}
  c_t = \kappa_t(\Mpool_t), \qquad a_t \sim \pi(\cdot \mid c_t).
\end{equation}
Actions come from three sets, $\Acts = \Agui \cup \Alook \cup \Acode$. GUI actions send input or change windows (click, type, press a key, scroll, drag, move the cursor, launch, raise or close a window), and $\Ain \subset \Agui$ are the input actions click, type, key press, scroll and drag. Look actions only perceive (list windows, capture a window, zoom). Code actions are shell commands, file tools, file inspection, reading a skill and declaring the task infeasible. GUI and look actions together are the desktop actions. The model may also answer without calling a tool, written $a_t = \varnothing$, which leaves the machine as it is. An admitted action is executed and returns a reply $o_t = (\sigma_t, y_t)$ with content $y_t$ and a mark $\sigma_t \in \{\checkmark, \times, -\}$: desktop actions report success or failure, and all other replies carry no mark. A refused action never reaches the machine:
\begin{equation}\label{eq:env}
  (s_{t+1}, o_t) =
  \begin{cases}
    \big(\mathrm{Env}(s_t, a_t),\ \omega(a_t, s_{t+1};\, z_t)\big) & \chi_t(a_t) = \textsf{allow},\\
    \big(s_t,\ (-, \text{refusal})\big) & \chi_t(a_t) = \textsf{refuse},
  \end{cases}
\end{equation}
where $z_t$ is a state the harness keeps for itself. The model $\pi$ is never trained or changed. Everything the agent achieves beyond the bare model comes from the harness configuration
\begin{equation}\label{eq:theta}
  \Theta = (\kappa_t,\ \beta,\ \chi_t,\ \omega),
\end{equation}
made of context construction $\kappa_t$, a read-only probing pass $\beta$ that we call the gate, which runs once before the first step and decides whether the run starts, $f \in \{0,1\}$, action admission $\chi_t$ and the reply function $\omega$. Let $g_t$ be the number of GUI actions among $a_1, \dots, a_{t-1}$. If $f = 0$ the run does not start. Otherwise it starts from $s_1 = s_0$ and ends at the first step where the model answers without a tool call, or right after the GUI action that uses up the budget $\Bgui$:
\begin{equation}\label{eq:stop}
  T = \min\{\, t : a_t = \varnothing \;\lor\; g_{t+1} \ge \Bgui \,\}.
\end{equation}
The task is scored once, by a function $r(s_{T+1}, d)$ that the agent never sees, of the state the run leaves behind and of whether the task was declared infeasible, $d \in \{0,1\}$, by the gate ($f = 0$, in which case the state is $s_0$) or by the agent. Two constraints hold throughout. The context holds at most $w$ screenshots, and the gate is restricted to read-only tools, with its probe shell instructed not to change anything.

The three challenges correspond to parts of $\Theta$. State and Context Management chooses $\kappa_t$ and $\chi_t$ during the run. Probing Before Acting defines $\beta$, which sets $f$ and the task-wide facts that $\kappa_t$ places in every context. Verifying After Acting defines $\omega$. Algorithm~\ref{alg:loop} shows where each part enters the loop.

\begin{algorithm}[t]
\caption{The harness loop. Comments name the mechanism that defines each step.}\label{alg:loop}
\small
\begin{algorithmic}[1]
\Require instruction $q$, machine state $s_0$, system prompt $P$, skill library $\Lib$, budget $\Bgui$
\State $E \gets \mathrm{snapshot}(s_0)$
\State $(G, \Xi, F, v, \eta) \gets \beta(q, E;\, \Lib, \Aro)$ \Comment{probing before acting}
\If{$v = \textsf{infeasible} \wedge F \ne \varnothing \wedge \eta \ne \varnothing$} \Return infeasible \Comment{$f = 0$}
\EndIf
\State $K \gets \bigoplus_{\gamma \in G} \gamma$;\ \ $b \gets h_v \oplus \bar b$;\ \ $s_1 \gets s_0$;\ \ $\Hist_1 \gets ()$;\ \ $z_1 \gets \mathbf{0}$
\For{$t = 1, 2, \dots$}
  \State $n_t \gets \nu(z_t;\, \bar b)$;\ \ $c_t \gets \kappa_t(\Mpool_t)$ \Comment{state and context management}
  \State $a_t \sim \pi(\cdot \mid c_t)$;\ \ \textbf{if} $a_t = \varnothing$ \textbf{then} $s_{t+1} \gets s_t$;\ \textbf{break}
  \If{$\chi_t(a_t) = \textsf{refuse}$} $s_{t+1} \gets s_t$;\ \ $o_t \gets (-, \text{refusal})$
  \Else\ $s_{t+1} \gets \mathrm{Env}(s_t, a_t)$;\ \ $o_t \gets \omega(a_t, s_{t+1};\, z_t)$ \Comment{verifying after acting}
  \EndIf
  \State $\Hist_{t+1} \gets \Hist_t \circ (a_t, o_t)$;\ \ $z_{t+1} \gets \mathcal{U}(z_t;\, a_t, o_t, n_t)$
  \State \textbf{if} $g_{t+1} \ge \Bgui$ \textbf{then break}
\EndFor
\end{algorithmic}
\end{algorithm}

\subsection{State and Context Management}\label{sec:state}

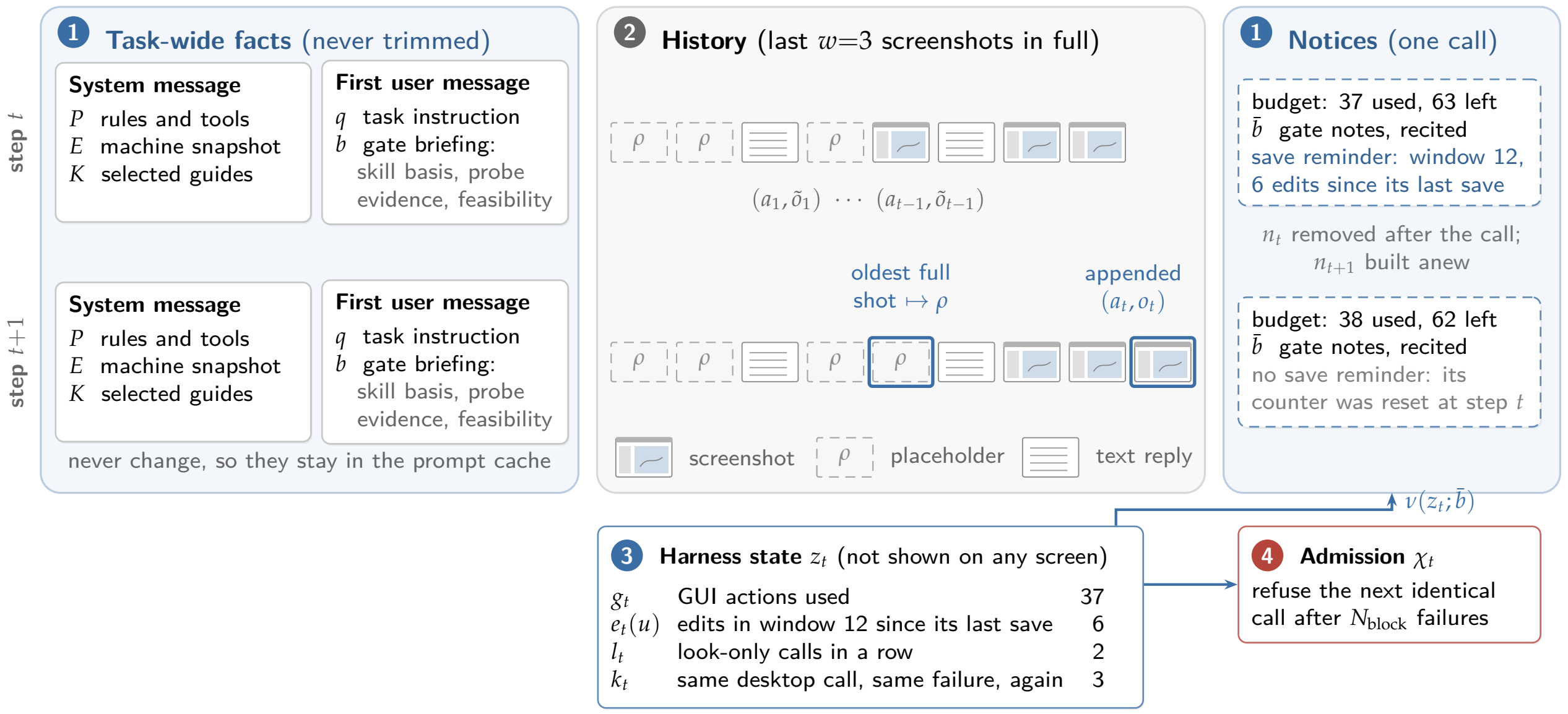
\begin{figure}[t]
  \centering
  \resizebox{\linewidth}{!}{\begin{tikzpicture}[fig]
  \fill[panel=ctxc]  (0,1.6)     rectangle (5.75,-3.6);
  \fill[panel=grayc] (5.95,1.6)  rectangle (12.45,-3.6);
  \fill[panel=ctxc]  (12.65,1.6) rectangle (16.25,-3.6);
  \node[badge=ctxc] at (0.35,1.33) {1};
  \node[phead=ctxc]  at (0.55,1.5)  {Task-wide facts \textmd{(never trimmed)}};
  \node[badge=black!60] at (6.3,1.33) {2};
  \node[phead=black] at (6.5,1.5)   {History \textmd{(last $w{=}3$ screenshots in full)}};
  \node[badge=ctxc] at (13.0,1.33) {1};
  \node[phead=ctxc]  at (13.2,1.5)  {Notices \textmd{(one call)}};

  \foreach \y/\s/\cap in {0.15/a/{step $t$}, -2.2/b/{step $t{+}1$}} {
    \node[lbl, rotate=90, font=\sffamily\figsc\bfseries] at (-0.25,\y) {\cap};
    \node[card, anchor=west, text width=2.45cm, minimum height=1.5cm] (sys\s) at (0.15,\y)
      {\textbf{System message}\\[2pt]$P$\ \ rules and tools\\$E$\ \ machine snapshot\\$K$\ \ selected guides\\\phantom{x}};
    \node[card, anchor=west, text width=2.35cm, minimum height=1.5cm] (usr\s) at (3.0,\y)
      {\textbf{First user message}\\[2pt]$q$\ \ task instruction\\$b$\ \ gate briefing:\\\textcolor{black!60}{\hspace*{0.9em}skill basis, probe}\\\textcolor{black!60}{\hspace*{0.9em}evidence, feasibility}};
  }
  \node[lbl] at (2.875,-3.27) {never change, so they stay in the prompt cache};

  \foreach \k/\p in {1/wingone,2/wingone,3/txtreply,4/wingone,5/win,6/txtreply,7/win,8/win} {
    \pic at ({6.4+(\k-1)*0.7},0.15) {\p};
  }
  \foreach \k/\p in {1/wingone,2/wingone,3/txtreply,4/wingone,5/wingone,6/txtreply,7/win,8/win,9/win} {
    \pic at ({6.4+(\k-1)*0.7},-2.2) {\p};
  }
  \foreach \k in {5,9}
    \draw[ctxc, line width=1pt, rounded corners=1pt] ({6.4+(\k-1)*0.7-0.34},-2.46) rectangle ({6.4+(\k-1)*0.7+0.34},-1.94);
  \node[lbl, text=ctxc, anchor=south] at ({6.4+4*0.7},-1.82) {oldest full\\shot $\mapsto\rho$};
  \node[lbl, text=ctxc, anchor=south east] at ({6.4+8*0.7+0.34},-1.82) {appended\\$(a_t,o_t)$};
  \node[lbl, anchor=north] at ({6.4+3.5*0.7},-0.2) {$(a_1,\tilde o_1)\ \cdots\ (a_{t-1},\tilde o_{t-1})$};
  \pic at (6.45,-3.22) {win};       \node[lbl, anchor=west] at (6.8,-3.22)  {screenshot};
  \pic at (8.6,-3.22) {wingone};    \node[lbl, anchor=west] at (8.95,-3.22) {placeholder};
  \pic at (10.8,-3.22) {txtreply};  \node[lbl, anchor=west] at (11.15,-3.22) {text reply};

  \node[card, anchor=west, draw=ctxc!80, densely dashed, text width=2.95cm, minimum height=1.35cm] (na) at (12.8,0.15)
    {budget: 37 used, 63 left\\$\bar b$\ \ gate notes, recited\\\textcolor{ctxc!90!black}{save reminder: window 12,}\\\textcolor{ctxc!90!black}{6 edits since its last save}};
  \node[lbl, text=black!55] at (14.45,-1.0) {$n_t$ removed after the call;\\$n_{t+1}$ built anew};
  \node[card, anchor=west, draw=ctxc!80, densely dashed, text width=2.95cm, minimum height=1.35cm] (nb) at (12.8,-2.2)
    {budget: 38 used, 62 left\\$\bar b$\ \ gate notes, recited\\\textcolor{black!50}{no save reminder: its}\\\textcolor{black!50}{counter was reset at step $t$}};

  \node[card, anchor=north west, draw=ctxc!80, text width=5.55cm] (z) at (5.95,-3.95) {%
    \tikz[baseline=(B.base)]\node[badge=ctxc](B){3};\ \ \textbf{Harness state $z_t$} \textmd{(not shown on any screen)}\\[3pt]
    \begin{tabular}{@{}l@{\ \ }l@{\ \ }r@{}}
      $g_t$    & GUI actions used                        & 37\\
      $e_t(u)$ & edits in window 12 since its last save  & 6\\
      $l_t$    & look-only calls in a row                & 2\\
      $k_t$    & same desktop call, same failure, again   & 3
    \end{tabular}};
  \node[card, anchor=north west, draw=failc!80, text width=2.95cm] (chi) at (12.8,-3.95) {%
    \tikz[baseline=(B.base)]\node[badge=failc](B){4};\ \ \textbf{Admission $\chi_t$}\\[2pt]refuse the next identical call after $\Nblock$ failures};
  \draw[arr, ctxc] (z.east |- chi.west) -- (chi.west);
  \draw[arr, ctxc] ([xshift=-0.3cm]z.north east) -- ++(0,0.17) -| (14.45,-3.6)
    node[pos=0.75, right, lbl, text=ctxc] {$\nu(z_t;\bar b)$};
\end{tikzpicture}}
  \caption{\textbf{Context construction and harness state.} The context sent to the model at two consecutive steps of a task, with illustrative values. Task-wide facts sit in the system and first user messages and never change. In the history only the last $w = 3$ screenshots stay in full: at step $t{+}1$ the new step is appended and the oldest full screenshot becomes the placeholder $\rho$. Notices are built from the harness state $z_t$ and the briefing, sent with one call and removed when it returns. The same state lets admission $\chi_t$ refuse a call that keeps failing the same way. Badges: 1 persistent facts and recitation, 2 bounded perception, 3 progress state, 4 repeated failures.}
  \label{fig:context}
\end{figure}

Context drifts in two ways over a long task. Screenshots accumulate until they crowd out what the task depends on, and the progress that matters most, such as the steps left or the work not yet saved, never appears on screen. We therefore choose $\kappa_t$ to keep perception within the bound $w$ while two kinds of information are never lost: facts the whole task depends on, and progress the model cannot observe. Figure~\ref{fig:context} shows the resulting context at two consecutive steps.

Formally, the information pool holds the system prompt $P$, the snapshot $E$ of the machine taken once before the task, the selected application skills $K$, the instruction $q$, the gate's briefing $b$, the history $\Hist_t = ((a_1, o_1), \dots, (a_{t-1}, o_{t-1}))$ and the harness state $z_t$:
\begin{equation}\label{eq:pool}
  \Mpool_t = \{P, E, K, q, b\} \cup \Hist_t \cup \{z_t\}.
\end{equation}
Context construction places the task-wide facts first, the history after them, and last the notices $n_t = \nu(z_t;\, \bar b)$ that the harness builds from its state and the briefing:
\begin{equation}\label{eq:kappa}
  \kappa_t(\Mpool_t) = \big[\, P \oplus E \oplus K,\;\; q \oplus b,\;\; (a_1, \tilde{o}_1), \dots, (a_{t-1}, \tilde{o}_{t-1}),\;\; n_t \,\big].
\end{equation}

\paragraph{Persistent facts and recitation.} The snapshot $E$ and the skills $K$ are part of the system message and the briefing $b$ is part of the first user message, so trimming never reaches them. Being present is not the same as being attended to, especially in the middle of a long context, and a long run tends to drift from its goal. Following the practice of reciting key task information at the end of the context~\citep{ji2025manus}, the harness repeats the body $\bar b$ of the briefing inside $n_t$, after the latest observation, before every model call. The notices are added for a single call and removed once it returns, so the history grows only by actions and replies.

\paragraph{Bounded perception.} Only perception is trimmed. Let $I_t$ be the indices of the last $w$ replies that carry a screenshot and $L_t$ the indices of the last $w$ replies of window captures and zooms. An older screenshot is replaced by a fixed placeholder $\rho$, which tells the model that the image was removed and that the screen may have changed since, and the text of an older capture or zoom reply is cut to a short preview $\phi(\cdot)$, with its full text saved to the workspace:
\begin{equation}\label{eq:window}
  \tilde{o}_j = (\sigma_j, \tilde{y}_j), \qquad
  \tilde{y}_j = y_j \;\text{ with }\;
  \begin{cases}
    \mathrm{shot} \mapsto \rho & j \notin I_t,\\
    \mathrm{text} \mapsto \phi(\mathrm{text}) & a_j \text{ a capture or zoom},\ j \notin L_t.
  \end{cases}
\end{equation}
Every other reply, such as shell output, file contents, window lists and inspection results, stays in full, and the history is never summarised. We use $w = 3$ so that a task moving between two windows keeps the latest frame of each in view. Because older replies are rewritten in place, the history after the first rewritten reply changes from step to step, and only the two leading messages are guaranteed to stay cached.

\paragraph{Progress state.} Rather than trusting the model to keep count, the harness keeps a state of what no observation shows:
\begin{equation}\label{eq:state}
  z_t = (g_t,\ \mathbf{e}_t,\ l_t,\ \mathbf{k}_t), \qquad \mathbf{e}_t = (e_t(u))_u,\quad \mathbf{k}_t = (k_t(\mathrm{id}, y))_{\mathrm{id}, y},
\end{equation}
where $g_t$ counts GUI actions so far, $e_t(u)$ counts edit actions (clicks, typing, key presses and drags, but not scrolls) addressed to window $u$ since its last save shortcut, $l_t$ counts look actions in a row, and $k_t(\mathrm{id}, y)$ counts identical failures of a desktop call. A save shortcut sets $e_t(u)$ to $0$. A save made through a menu is not seen. Any non-look action sets $l_t$ to $0$. The notices state the budget, recite the briefing, and add a save or a look-only reminder when a counter reaches its threshold:
\begin{equation}\label{eq:notice}
  n_t = \nu(z_t;\, \bar b) = \big[\, (g_t, \Bgui - g_t),\;\; \bar b,\;\; \mathrm{save}(u^\ast) \text{ if } e_t(u^\ast) \ge \Nsave,\;\; \mathrm{look} \text{ if } l_t \ge \Nlook \,\big],
\end{equation}
where $u^\ast = \arg\max_u e_t(u)$. The update $z_{t+1} = \mathcal{U}(z_t;\, a_t, o_t, n_t)$ first resets every counter that $n_t$ reported and then counts $a_t$, so a reminder the model ignores returns only after another $\Nsave$ edits or $\Nlook$ looks. The harness never saves on the model's behalf, because the same save shortcut does different things in different applications and dialogs. The budget in view lets the agent save and export its result before it runs out instead of refining until the last step.

\paragraph{Repeated failures.} For desktop actions, $k_t(\mathrm{id}, y)$ counts how often the call identity $\mathrm{id}$, a tool with its exact arguments, has returned the same non-success reply $y$ in the run. A success of the call clears its entries. When a call returns the same non-success reply for the $\Nadv$-th time, advice to change approach is appended to that reply and stays in the history, which is why $\omega$ reads $z_t$. Once that count has reached $\Nblock$, admission refuses the next identical call before it reaches the machine:
\begin{equation}\label{eq:admit}
  \chi_t(a_t) =
  \begin{cases}
    \textsf{refuse} & \max_y k_t(\mathrm{id}(a_t), y) \ge \Nblock,\\
    \textsf{allow} & \text{otherwise}.
  \end{cases}
\end{equation}
We use $\Nsave = 6$, $\Nlook = 8$, $\Nadv = 3$ and $\Nblock = 6$.

\subsection{Probing Before Acting}\label{sec:probe}

\begin{figure}[t]
  \centering
  \resizebox{\linewidth}{!}{\begin{tikzpicture}[fig]
  \node[banner, anchor=north west, text width=15.07cm] (task) at (0,2.35)
    {\textbf{Task $q$:}\ \ In the image editor, crop \texttt{scan.png} to the signature and export it as \texttt{signature.png}.};

  \fill[panel=prbc] (0,1.45) rectangle (6.95,-3.2);
  \node[phead=prbc] at (0.15,1.35) {Gate $\beta$ \textmd{(read-only, before the first action)}};
  \node[card, anchor=north west, text width=6.1cm] (g1) at (0.2,0.9) {%
    \tikz[baseline=(B.base)]\node[badge=prbc](B){1};\ \textbf{Load knowledge}\hfill guides $G$\\[1pt]
    {\ttfamily select\_skills("image-editor")}};
  \node[card, anchor=north west, text width=6.1cm] (g2) at ([yshift=-0.15cm]g1.south west) {%
    \tikz[baseline=(B.base)]\node[badge=prbc](B){2};\ \textbf{Probe the environment}\hfill findings $F$\\[1pt]
    {\ttfamily \$ ls -l \textasciitilde/Pictures/scan.png}\hfill\textcolor{prbc!80!black}{found, 2480$\times$3508}\\
    {\ttfamily \$ gimp {-}{-}version}\hfill\textcolor{prbc!80!black}{2.10.38, crop and export built in}};
  \node[card, anchor=north west, text width=6.1cm] (g3) at ([yshift=-0.15cm]g2.south west) {%
    \tikz[baseline=(B.base)]\node[badge=prbc](B){3};\ \textbf{Decide feasibility}\hfill verdict $v$, reason $\eta$\\[1pt]
    feasible: the file exists; the editor crops and exports};
  \node[lbl, anchor=north west, align=flush left, text width=6.6cm] at ([yshift=-0.08cm]g3.south west)
    {read-only tools $\Aro$, incl.\ a hidden probe shell; $\le N_\beta$ calls};
  \draw[arr, prbc] (g1) -- (g2);
  \draw[arr, prbc] (g2) -- (g3);

  \node[diamond, aspect=1.45, draw=black!45, fill=white, inner sep=0.5pt, align=flush center, font=\sffamily\figsc]
    (d) at (8.45,-0.35) {infeasible\\with evidence?};
  \draw[arr] (g3.east) -- ++(0.3,0) |- (d.west);
  \node[card, draw=failc!80, anchor=north, text width=2.65cm] (stop) at (8.45,-1.4)
    {\textbf{End the task} ($f{=}0$)\\[1pt]\textcolor{black!55}{e.g.\ the file is missing}\\$\Rightarrow$ infeasible, with evidence};
  \draw[arr, failc] (d.south) -- node[lbl, right] {yes} (stop.north);

  \fill[panel=ctxc] (9.95,1.45) rectangle (15.35,-3.2);
  \node[phead=ctxc] at (10.1,1.35) {Task-wide facts \textmd{before step 1}};
  \draw[arr] (d.east) -- node[lbl, above] {no} (9.95,-0.35);
  \node[card, anchor=north west, text width=4.75cm] (es) at (10.1,0.9) {%
    \textbf{System message}\\[1pt]$P$ rules,\ \ $E$ snapshot,\ \ $K$ selected guides};
  \node[card, anchor=north west, text width=4.75cm] (eu) at ([yshift=-0.12cm]es.south west) {%
    \textbf{First user message}\ \ task $q$ and briefing $b$\\[2pt]
    \textcolor{prbc!80!black}{\textbf{Skill basis}}\ \ \textit{Image editor $\rangle$ Editing rules:} When the task is done in the image editor, use the editor's own tools. Do not edit the image with other programs.\\[1pt]
    \textcolor{prbc!80!black}{\textbf{Probe evidence}}\ \ scan.png found (2480$\times$3508); editor 2.10.38\\[1pt]
    \textcolor{prbc!80!black}{\textbf{Feasibility}}\ \ feasible};

  \begin{scope}[yshift=-0.2cm]
  \fill[panel=grayc] (0,-3.35) rectangle (15.35,-5.55);
  \node[badge=prbc] at (0.35,-3.6) {4};
  \node[phead=black] at (0.55,-3.42) {Expand quotes\ \ \textmd{the gate's quote only locates the rule}};
  \node[card, anchor=north west, text width=3.6cm] (qq) at (0.2,-3.9)
    {quote from the gate\\[1pt]\textit{``\dots\ do not edit the image with other programs''}};
  \node[card, anchor=north west, text width=6.6cm, draw=prbc!80] (uu) at (5.0,-3.9)
    {unit $x$ of the guide, with heading path $H(x)$\\[1pt]
     \textit{Image editor $\rangle$ Editing rules:} \textbf{When the task is done in the image editor,} use the editor's own tools. \textcolor{prbc!80!black}{\textbf{Do not edit the image with other programs.}}};
  \draw[arr] (qq.east) -- node[lbl, above] {match} (uu.west |- qq.east);
  \node[lbl, anchor=west, text width=3.0cm, align=flush left] at (12.05,-4.75)
    {the condition in bold now travels with the rule; a quote that matches nothing is dropped};
  \draw[arr, prbc, densely dashed] (11.2,-3.9) -- node[lbl, right, pos=0.22, text=prbc] {becomes the skill basis} (eu.south -| 11.2,0);
  \end{scope}
\end{tikzpicture}}
  \caption{\textbf{Probing before acting}, on an example task. The gate loads the skill the task needs, probes the named file and the installed editor with read-only tools, and decides feasibility. An infeasible verdict ends the task only when it comes with findings, such as a missing file, and a reason. Otherwise the skills enter the executor's system prompt and the briefing its first user message. Bottom: a rule the gate quotes is replaced by the paragraph of the skill that contains it, with its heading path, so the condition in bold travels with the rule. Badges 1--4 match the paragraph headings in the text.}
  \label{fig:gate}
\end{figure}
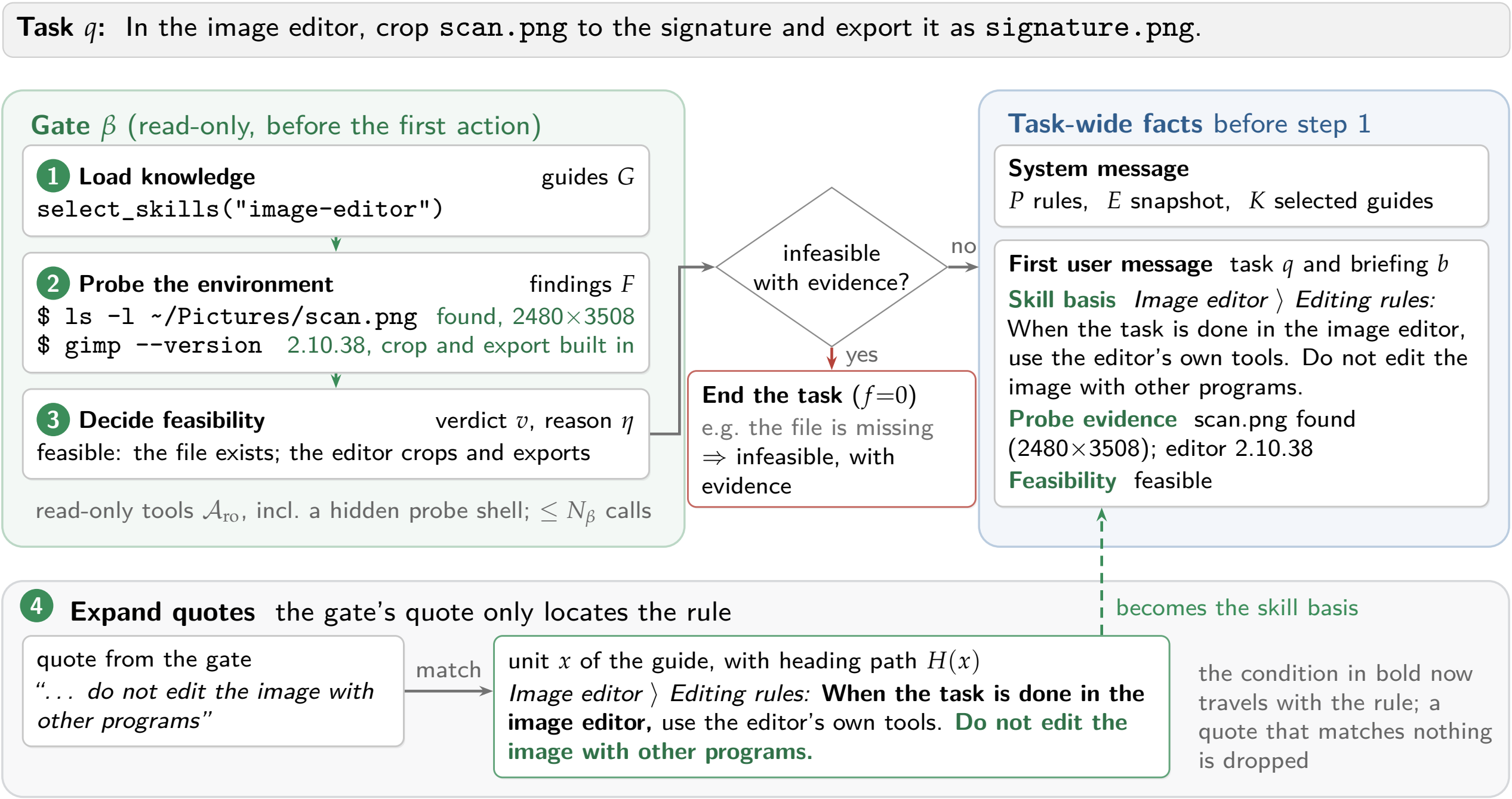

To close the knowledge gap, the agent's picture of the task has to be checked before it acts. Knowledge offered on demand is read only when the agent knows it is missing, so it goes unread exactly when it matters: when the agent does not know what it lacks. We therefore gather it before the first action, in the gate. The gate decides $f$, whether the run starts, and sets the skills $K$ and the briefing $b$ that every later context carries. State and Context Management keeps these facts in view, and the gate determines what they are and checks them against evidence. Figure~\ref{fig:gate} follows the gate on an example task.

Formally, the gate is a probing pass $\beta$. It may use only the read-only actions $\Aro$: the look actions $\Alook$, file reads, searches, file-type checks and inspection, together with a probe shell that only the gate has. It makes at most $N_\beta$ model calls, plus one final request to answer if it ends without a verdict. It draws on a library $\Lib$ of application skills, each skill $\gamma$ a sequence of units $x$ (its paragraphs, with each top-level list item counted as one), with $H(x)$ the heading path above unit $x$. It returns the skills it loaded $G \subseteq \Lib$, its quotes from them $\Xi = (\xi_1, \dots, \xi_m)$, its findings $F$ (a probe-evidence section), a verdict $v$ and a reason $\eta$:
\begin{equation}\label{eq:beta}
  (G, \Xi, F, v, \eta) = \beta(q, E;\, \Lib, \Aro), \qquad v \in \{\textsf{feasible}, \textsf{infeasible}\}.
\end{equation}

\paragraph{Load knowledge.} The full text of the loaded skills goes into the executor's system prompt, $K = \bigoplus_{\gamma \in G} \gamma$, and the executor can still open any other skill during the run. Skills state facts about the software, such as what a menu does and which features the installed version has, along with working rules for that application. Whether a task is feasible and what counts as done are decided only by rules in the system prompt, so a skill never overrides those judgements.

\paragraph{Probe the environment.} Starting from the snapshot $E$ (operating system, installed application versions and editor extensions, open windows, the files on the desktop and in the home directory, and the current date), the gate inspects the files, settings and applications the task names, and records in $F$ what it saw and what it could not confirm. The probe shell runs outside the session the executor will use, so its commands leave no trace there.

\paragraph{Decide feasibility.} People do ask for things their software cannot do, and the useful response is to say so rather than to produce something that only looks right. The gate and the executor share one rule for this: a task asks either for a result or for a capability, and if it asks for a capability the application lacks, building that capability with a script or another program does not complete it. A verdict drawn from appearances alone can mistake a feasible task for an infeasible one, and ending a feasible task forfeits it. The run is therefore skipped only on an infeasible verdict that comes with findings and a reason:
\begin{equation}\label{eq:feasible}
  f = 1 - \mathbf{1}\big[\, v = \textsf{infeasible} \,\wedge\, F \ne \varnothing \,\wedge\, \eta \ne \varnothing \,\big].
\end{equation}
Any other verdict is passed on, and an executor told that a task looks infeasible first tries the application's own way once. A gate that ends without a verdict is asked once more to answer from what it found. If it still gives none, its notes are passed on as feasible.

\paragraph{Expand quotes.} A rule quoted out of its skill tends to lose its condition. ``Do not edit the image with other programs'', quoted without the ``when the task is done in the image editor'' before it, will be applied by an executor even where the condition does not hold. The harness therefore uses each quote only to locate a rule. A quote $\xi_i$ from skill $\gamma_i$ is split at ellipses into fragments of at least $\ell_{\min}$ characters, $\mathrm{frag}(\xi_i)$, and replaced by every paragraph of $\gamma_i$ that contains one of them, each prefixed with its heading path. Writing $\zeta \sqsubseteq x$ when $\zeta$ is a substring of $x$ after both are normalised (case, spaces and markup),
\begin{equation}\label{eq:expand}
  \mathrm{expand}(\xi_i) = \bigoplus_{x \in \gamma_i \,:\, \exists \zeta \in \mathrm{frag}(\xi_i),\ \zeta \sqsubseteq x} H(x) \oplus x.
\end{equation}
A quote that matches no paragraph is dropped, which also keeps the gate from inventing rules. The briefing is a lead $h_v$ that depends on the verdict, followed by a body $\bar b$ in three labelled parts: \emph{Skill basis} with the expanded rules, \emph{Probe evidence} with the findings, and \emph{Feasibility} with the verdict and its reason:
\begin{equation}\label{eq:init}
  b = h_v \oplus \bar b, \qquad
  \bar b = \big(\textstyle\bigoplus_i \mathrm{expand}(\xi_i),\; F,\; (v, \eta)\big).
\end{equation}
We use $N_\beta = 20$ and $\ell_{\min} = 15$. If the gate itself fails to run, $f = 1$, $b$ is empty and $K$ holds the skills of the applications that $q$ names or that are open in $E$ (the general system skill if none).

\subsection{Verifying After Acting}\label{sec:verify}

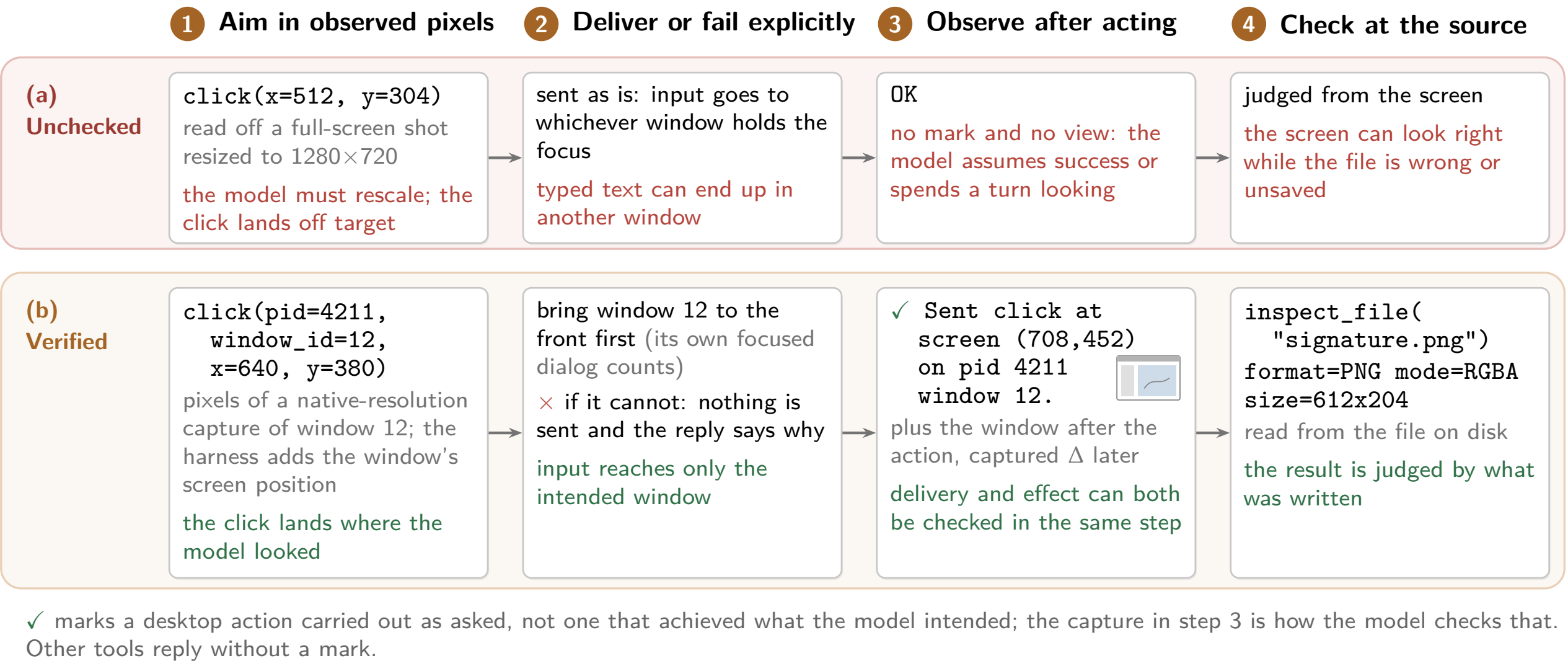
\begin{figure}[t]
  \centering
  \resizebox{\linewidth}{!}{\begin{tikzpicture}[fig,
    cell/.style={card, anchor=north west, inner sep=4pt},
    stepn/.style={circle, fill=verc!85!black, text=white, inner sep=0pt, minimum size=0.36cm, font=\sffamily\figsc\bfseries},
  ]
  \newcommand{\ok}{\textcolor{prbc!85!black}{\ensuremath{\checkmark}}}
  \newcommand{\no}{\textcolor{failc}{\ensuremath{\times}}}
  \newcommand{\gr}[1]{\textcolor{black!55}{#1}}
  \newcommand{\bad}[1]{\textcolor{failc}{#1}}
  \newcommand{\good}[1]{\textcolor{prbc!80!black}{#1}}
  \newcommand{\cb}[2]{\parbox[c][#1][t]{3.05cm}{\raggedright #2}}
  \def\xa{1.75} \def\xb{5.45} \def\xc{9.15} \def\xd{12.85}

  \foreach \x/\n/\t in {\xa/1/{Aim in observed pixels}, \xb/2/{Deliver or fail explicitly}, \xc/3/{Observe after acting}, \xd/4/{Check at the source}} {
    \node[stepn] (h\n) at (\x+0.2,2.3) {\n};
    \node[anchor=west, font=\sffamily\figfn\bfseries] at (h\n.east) {\t};
  }

  \fill[panel=failc] (0,1.95) rectangle (16.35,-0.05);
  \node[phead=failc, align=flush left, font=\sffamily\figsc\bfseries] at (0.12,1.8) {(a)\\[1pt]Unchecked};
  \node[cell] (a1) at (\xa,1.8) {\cb{1.5cm}{\texttt{click(x=512, y=304)}\\[1pt]\gr{read off a full-screen shot resized to 1280$\times$720}\\[3pt]\bad{the model must rescale; the click lands off target}}};
  \node[cell] (a2) at (\xb,1.8) {\cb{1.5cm}{sent as is: input goes to whichever window holds the focus\\[3pt]\bad{typed text can end up in another window}}};
  \node[cell] (a3) at (\xc,1.8) {\cb{1.5cm}{\texttt{OK}\\[3pt]\bad{no mark and no view: the model assumes success or spends a turn looking}}};
  \node[cell] (a4) at (\xd,1.8) {\cb{1.5cm}{judged from the screen\\[3pt]\bad{the screen can look right while the file is wrong or unsaved}}};

  \fill[panel=verc] (0,-0.3) rectangle (16.35,-3.6);
  \node[phead=verc, align=flush left, font=\sffamily\figsc\bfseries] at (0.12,-0.45) {(b)\\[1pt]Verified};
  \node[cell] (b1) at (\xa,-0.45) {\cb{2.75cm}{\texttt{click(pid=4211,}\\\texttt{\ \ window\_id=12,}\\\texttt{\ \ x=640, y=380)}\\[1pt]\gr{pixels of a native-resolution capture of window 12; the harness adds the window's screen position}\\[3pt]\good{the click lands where the model looked}}};
  \node[cell] (b2) at (\xb,-0.45) {\cb{2.75cm}{bring window 12 to the front first \gr{(its own focused dialog counts)}\\[2pt]\no\ if it cannot: nothing is sent and the reply says why\\[3pt]\good{input reaches only the intended window}}};
  \node[cell] (b3) at (\xc,-0.45) {\cb{2.75cm}{\texttt{\ok\ Sent click at}\\\texttt{\ \ screen (708,452)}\\\texttt{\ \ on pid 4211}\\\texttt{\ \ window 12.}\\[1pt]\gr{plus the window after the action, captured $\Delta$ later}\\[3pt]\good{delivery and effect can both be checked in the same step}}};
  \pic[scale=1.1] at ([xshift=-0.5cm,yshift=-0.95cm]b3.north east) {win};
  \node[cell] (b4) at (\xd,-0.45) {\cb{2.75cm}{\texttt{inspect\_file(}\\\texttt{\ \ "signature.png")}\\[1pt]\texttt{format=PNG mode=RGBA}\\\texttt{size=612x204}\\[1pt]\gr{read from the file on disk}\\[3pt]\good{the result is judged by what was written}}};

  \foreach \l in {a,b} {
    \draw[arr] (\l1.east) -- (\l2.west);
    \draw[arr] (\l2.east) -- (\l3.west);
    \draw[arr] (\l3.east) -- (\l4.west);
  }
  \node[lbl, anchor=north west, align=flush left, text width=16cm] at (0.12,-3.7)
    {\ok\ marks a desktop action carried out as asked, not one that achieved what the model intended; the capture in step 3 is how the model checks that. Other tools reply without a mark.};
\end{tikzpicture}}
  \caption{\textbf{Verifying after acting}: an input action and the later check of its saved result. (a) Without checks, the input is aimed at a resized screenshot, goes to whichever window holds the focus and returns a bare acknowledgement, and the saved result is judged from the screen. Each step can fail without the model noticing. (b) Our input is aimed in the pixels of a native-resolution capture of its window, is sent only if that window can be brought to the front, and returns a mark with a fresh capture of the window, so an action that failed or went wrong shows up in the step it happens, before later steps build on it. The saved file is then read back from disk.}
  \label{fig:feedback}
\end{figure}

To stop errors from accumulating, the agent has to verify what each action did before it builds on it. An outcome can be verified only if it is checkable: the action must name its target in the frame the model sees, and must either reach that target or report that it did not. We therefore define $\omega$ so that each input action returns a checkable outcome. Its target is given in the pixels of a capture of its window. It is delivered only if that window can be brought to the front, and otherwise reports a failure. On success its reply carries a mark and a fresh capture of the window. A saved result can then be checked at its source. Figure~\ref{fig:feedback} follows an input action and the later check of its result through these four steps, with and without the checks. The harness adds no model call: it makes each outcome checkable, and the model does the verifying.

\paragraph{Aim in observed pixels.} An input action names its target window $u_t$ by its process and, optionally, a window id. A capture is taken at the native resolution of the window it shows, and the action gives its coordinates in the pixel frame of that capture, and the harness adds the window's position on screen. The position the model reads off the image is therefore the position the input uses. A downscaled screenshot instead makes the model rescale every coordinate itself, an error that is systematic and silent.

\paragraph{Deliver or fail explicitly.} Input is real foreground input, so an input action is sent only if $u_t$ can be brought to the front. A dialog that belongs to $u_t$ and already holds the focus counts as $u_t$, so the harness does not pull the focus away from a dialog the task has opened. Otherwise nothing is sent and the reply is a failure:
\begin{equation}\label{eq:focus}
  \neg\, \mathrm{front}(u_t) \;\Rightarrow\; \sigma_t = \times,\quad s_{t+1} = s_t.
\end{equation}
An input that names no window is sent at screen coordinates, without this check and without the capture that follows a successful input.

\paragraph{Observe after acting.} The reply of a desktop action starts with its mark. A success mark records that the action was carried out as asked, for an input action that it was delivered at the stated screen position in the stated window. Whether it had the intended effect is for the model to judge. When an input action succeeds, its reply also carries a capture of $u_t$ taken a settle time $\Delta$ after the action, so the model sees the effect in the same step instead of spending a separate turn looking:
\begin{equation}\label{eq:omega}
  a_t \in \Ain \,\wedge\, \sigma_t = \checkmark \;\Rightarrow\; y_t \ni \mathrm{shot}(s_{t+1};\, u_t).
\end{equation}
Launching or raising a window and moving the cursor return text only. We use $\Delta = 0.8$\,s.

\paragraph{Check at the source.} What the screen shows is the application's rendering of a document, not the document. The agent can inspect a saved file with an ordinary code action, which reads attributes from the file itself, such as fonts, colours, number formats, cell values and layout, so a saved result can be judged by what was written to disk.

\section{Experiments}\label{sec:experiments}

\subsection{Experimental Setup}\label{sec:setup}

\paragraph{Benchmark.} We evaluate on OSWorld-Verified~\citep{xie2024osworld}, 361 tasks on a real Ubuntu desktop across ten domains: Chrome, GIMP, LibreOffice Calc, Impress and Writer, multi-application workflows, the operating system, Thunderbird, VLC and VS Code. Tasks, their setup and their graders are used unmodified.

\paragraph{Models and configuration.} The agent runs on Claude Opus~5 with every mechanism of Section~\ref{sec:method} enabled: the gate is binding, input actions return a fresh window capture, and the remaining GUI action budget is shown before each model call. The budget is 100 GUI actions per task. As for the agentic-framework entries on the leaderboard, only GUI actions count toward it. Looking at the screen, reading files and shell commands do not. The prompt, tools and model configuration are fixed across all tasks. The model runs at its default reasoning setting, adaptive thinking at effort ``high''.

\paragraph{Metric.} Success rate: the summed task score divided by the number of tasks, with each task graded by the benchmark's own checker on the state the agent leaves behind. Some checkers give partial credit. Prior results are taken from the official leaderboard and are not rerun.

\subsection{Results on OSWorld-Verified}\label{sec:results}
\paragraph{Results.} As shown in Table~\ref{tab:leaderboard}, our agent achieves \textbf{93.5\%} (337.71\,/\,361), \textbf{3.3 percentage points} above the strongest leaderboard entry, Intelligence-Indeed Agent (90.2\%), and \textbf{10.1 percentage points} above Anthropic's own agent for Claude Opus~5 (83.4\%).

\begin{table}[t]
  \centering
  \footnotesize
  \setlength{\tabcolsep}{3pt}
  \caption{Comparison with leading systems on OSWorld-Verified (100-step budget). Prior results are the top seven entries of the official leaderboard, and all use a 100-step budget and a single run. -- marks a model the entry does not name. Ours is not a leaderboard entry.}
  \label{tab:leaderboard}
  \begin{tabular}{c>{\raggedright}p{2.5cm}>{\raggedright}p{2.1cm}>{\raggedright}p{1.9cm}>{\raggedright}p{2.1cm}>{\raggedright}p{1.6cm}cc}
    \toprule
    Rank & Agent / Harness & Agent Org. & Model & Model Org. & Type & Date & Success (\%) \tabularnewline
    \midrule
    1 & Intelligence-Indeed Agent & Intelligence Indeed & -- & -- & Agentic framework & 2026-07-25 & 90.2 \tabularnewline
    2 & claude-fable-5[1m] & Anthropic & Claude Fable 5 & Anthropic & General model & 2026-08-01 & 86.0 \tabularnewline
    3 & Pointer Agent & Pointer & Claude Opus 4.7 & Anthropic & Agentic framework & 2026-05-21 & 83.6 \tabularnewline
    4 & claude-opus-5[1m] & Anthropic & Claude Opus 5 & Anthropic & General model & 2026-08-01 & 83.4$^\dagger$ \tabularnewline
    5 & Coasty CUA v1 & Coasty Team & -- & -- & General model & 2026-07-01 & 82.8 \tabularnewline
    6 & Pointer Agent & Pointer & Claude Sonnet 4.6 & Anthropic & Agentic framework & 2026-05-21 & 81.5 \tabularnewline
    7 & Muse Spark 1.1 & Meta Superintelligence Labs & Muse Spark 1.1 & Meta Superintelligence Labs & General model & 2026-07-09 & 80.7 \tabularnewline
    \midrule
    Ours & openJiuwen agent-core & openJiuwen & Claude Opus 5 & Anthropic & Agentic framework & -- & \textbf{93.5} \tabularnewline
    \bottomrule
  \end{tabular}

  \vspace{2pt}
  \raggedright\footnotesize $^\dagger$Over the 360 tasks it ran. The entry has 25 of the 26 GIMP tasks.
\end{table}

\paragraph{Domain-level analysis.} Table~\ref{tab:domains} breaks the results down by domain. The agent has the best or tied-best score in nine of the ten domains and solves every task in GIMP, LibreOffice Writer and OS. The exception is VLC, where it trails Intelligence-Indeed Agent by 1.3 points (16.67 against 16.89 in summed score). The largest margins over the strongest entry are in GIMP (+7.7), Thunderbird (+6.7), LibreOffice Impress (+6.3) and multi-application workflows (+4.7). The last is the largest domain, with 93 tasks.

\begin{table}[t]
  \centering
  \small
  \caption{Domain-level success rates (\%) on OSWorld-Verified. Bold indicates the best score in each row (ties included).}
  \label{tab:domains}
  \begin{tabular}{lcccc}
    \toprule
    Domain & Tasks & Anthropic Opus 5 & Intelligence-Indeed & Ours \tabularnewline
    \midrule
    Chrome              & 46  & 80.2             & 78.1           & \textbf{80.3}  \tabularnewline
    GIMP                & 26  & 88.0$^\dagger$   & 92.3           & \textbf{100.0} \tabularnewline
    LibreOffice Calc    & 47  & 95.7             & \textbf{97.9}  & \textbf{97.9}  \tabularnewline
    LibreOffice Impress & 47  & 80.6             & 91.4           & \textbf{97.7}  \tabularnewline
    LibreOffice Writer  & 23  & 95.7             & 95.7           & \textbf{100.0} \tabularnewline
    Multi-apps          & 93  & 76.8             & 84.7           & \textbf{89.5}  \tabularnewline
    OS                  & 24  & 79.2             & \textbf{100.0} & \textbf{100.0} \tabularnewline
    Thunderbird         & 15  & 73.3             & 86.7           & \textbf{93.3}  \tabularnewline
    VLC                 & 17  & 82.4             & \textbf{99.4}  & 98.1           \tabularnewline
    VS Code             & 23  & 91.3             & \textbf{95.7}  & \textbf{95.7}  \tabularnewline
    \midrule
    Overall             & 361 & 83.4             & 90.2           & \textbf{93.5}  \tabularnewline
    \bottomrule
  \end{tabular}

  \vspace{2pt}
  \footnotesize $^\dagger$Over the 25 GIMP tasks the entry ran.
\end{table}

\paragraph{Model-matched comparison.} Notably, our agent and Anthropic's run the same model, Claude Opus~5, under the same 100-step budget, so the gap between them reflects the harness rather than the model. The agent is ahead of Anthropic's in every domain, by the most in OS (+20.8), Thunderbird (+20.0), LibreOffice Impress (+17.0) and VLC (+15.7). The margin is smallest in Chrome (+0.1), which is also where most of the remaining loss lies: Chrome (9.06) and multi-application workflows (9.80) together account for 81\% of the 23.29 points of score the agent does not obtain.

The domains with the largest gains, such as OS, Thunderbird and Impress, are ones where a task's success depends on the exact state of an application or a file rather than on what the screen shows. This is consistent with checking saved results and tracking unsaved work, but it should be read as a plausible systems-level factor rather than a controlled causal attribution. More broadly, these are system-level comparisons rather than controlled estimates of each mechanism's contribution. Section~\ref{sec:ablation} removes each mechanism in turn.

\subsection{Ablation Study}\label{sec:ablation}

\paragraph{Setup.} To isolate each mechanism, we turn off one mechanism at a time and keep every other setting of Section~\ref{sec:setup} fixed. Each ablation runs once, with the same 100-step budget, on a subset of domains where its mechanism is exercised:
\begin{itemize}
  \item \textbf{Without Probing Before Acting}, on OS and GIMP (50 tasks). The two subsets are chosen due to the percentage of infeasible questions lying mostly with these two subsets, making it the two most suitable subsets of data to do this ablation study on. The gate is removed, so there is no environment probing, feasibility decision or briefing. The skills are not preloaded and the machine snapshot is not given.
  \item \textbf{Without State and Context Management}, on multi-application workflows (93 tasks). All five state signals are turned off: remaining budget, save reminder, look-only reminder, recitation of the briefing, and repeated-failure advice and refusal. We determined that the subset that would suffer the most from this change would be the multi-apps subset, and therefore we deem this the most suitable subset to run this ablation study on. We run this twice, keeping only the latest screenshot ($w = 1$) and keeping every screenshot ($w = \infty$). The full agent uses $w = 3$.
  \item \textbf{Without Verifying After Acting}, on LibreOffice Writer and GIMP (49 tasks). Three checks are removed: the capture of the window after each input action, the wait for the target window to become active together with the failure reply when it does not, and the success or failure mark on each reply. Input is still aimed in the pixels of a window capture, the target window is still raised before input, and a saved file can still be inspected with a code action. The two domains cover both places where a result can be checked: in GIMP an edit is judged on the canvas, and in Writer the result is a document that can also be checked at its source.
\end{itemize}
The full agent's scores are those of Table~\ref{tab:domains}. Where the per-step logs of a full-agent run with the same model and agent framework are available, we also compare how the agent works and what it costs (Table~\ref{tab:process}).

\begin{table}[t]
  \centering
  \small
  \setlength{\tabcolsep}{4pt}
  \caption{Ablation results on OSWorld-Verified (100-step budget, one run per configuration). $\Delta$ is the full agent's score minus the ablated score, and a positive value means the agent is worse without the mechanism. Percentage points are $\Delta$ divided by the number of tasks.}
  \label{tab:ablation}
  \begin{tabular}{llcccrr}
    \toprule
    Configuration & Subset & Tasks & Full & Ablated & $\Delta$ (score) & $\Delta$ (pp) \tabularnewline
    \midrule
    \multirow{3}{*}{w/o Probing Before Acting}
      & OS                & 24 & 24.00 & 22.00 & 2.00 & 8.3 \tabularnewline
      & GIMP              & 26 & 26.00 & 25.00 & 1.00 & 3.8 \tabularnewline
      & \textit{Total}    & 50 & 50.00 & 47.00 & \textbf{3.00} & \textbf{6.0} \tabularnewline
    \midrule
    \shortstack[l]{w/o State and Context\\Management, $w = 1$}
      & Multi-apps & 93 & 83.19 & 77.08 & \textbf{6.11} & \textbf{6.6} \tabularnewline
    \addlinespace
    \shortstack[l]{w/o State and Context\\Management, $w = \infty$}
      & Multi-apps & 93 & 83.19 & 78.08 & \textbf{5.11} & \textbf{5.5} \tabularnewline
    \midrule
    \multirow{3}{*}{w/o Verifying After Acting}
      & LibreOffice Writer & 23 & 23.00 & 22.00 & 1.00 & 4.3 \tabularnewline
      & GIMP               & 26 & 26.00 & 25.00 & 1.00 & 3.8 \tabularnewline
      & \textit{Total}     & 49 & 49.00 & 47.00 & \textbf{2.00} & \textbf{4.1} \tabularnewline
    \bottomrule
  \end{tabular}
\end{table}

\paragraph{Results.} As shown in Table~\ref{tab:ablation}, the score on each subset is lower with the mechanism removed, although for Verifying After Acting the lost points do not trace to a missing check (see below). The largest drop comes from State and Context Management: on the 93 multi-application tasks the agent loses \textbf{6.11 points} (6.6 percentage points) when only the latest screenshot is kept and \textbf{5.11 points} (5.5 percentage points) when every screenshot is kept. Removing Probing Before Acting costs \textbf{3 tasks} out of 50, two of them in OS, and removing Verifying After Acting costs \textbf{2 points} out of 49, one in each domain.

\paragraph{Probing Before Acting.} The ablation removes everything the agent is given before it starts: the gate, the skill rules the gate selects and injects, and the snapshot of the machine. The agent can still look up skills on demand and can still declare a task infeasible. We ran it on OS and GIMP, the two domains that hold most of the benchmark's infeasible tasks (50 tasks, 14 of them infeasible). The ablated agent solved 47. It declared all 14 infeasible tasks infeasible, so the failure that the feasibility rule of Section~\ref{sec:probe} guards against, building a missing capability with a script or another program instead of declaring the task infeasible, did not appear. The three tasks it lost are all feasible, and two are in OS, where the full configuration solved all 24 tasks. On \texttt{gimp/dbbf4b99} (convert a RAW image to JPEG using GIMP only), GIMP reported that no RAW loader was installed. The agent confirmed that neither darktable nor RawTherapee was present and declared the task infeasible, which is the other failure Section~\ref{sec:probe} describes: mistaking a feasible task, one that earlier runs had completed, for an infeasible one. The two OS tasks were lost for a different reason: each missed a convention that the gate's injected skills state. On \texttt{os/5812b315}, the home directory of a restricted SSH user had to be the allowed folder itself, and the agent made it a subfolder. On \texttt{os/5c1075ca}, ``the current directory'' meant the directory of the open terminal, and the agent ran the copy from the home directory instead. In both cases the full configuration followed the injected rule. On these tasks, then, most of what probing contributes comes from passing the right conventions to the executor rather than from deciding feasibility.

\paragraph{State and Context Management.} Both arms lose more than 5 points, and the loss does not go away in either direction of the screenshot window: keeping every screenshot ($w = \infty$) is only 1.0 point better than keeping one ($w = 1$). Multi-application tasks explain why both ends fail. They move content and actions from one window to the next: a value read in a spreadsheet is typed into a document, a file saved in one application is attached in another. With $w = 1$, the moment the agent looks at window $n{+}1$, the last frame of window $n$ is gone, so it has to carry what that window held, and what it had already done there, in its own reasoning. The snapshot of the source window that the next step depends on is no longer in view. With $w = \infty$ the opposite happens. The context fills with screenshots, most of them stale, that show windows and states that no longer exist. They can mislead the model, and they leave it to work out which of many frames reflects the current screen. Keeping the last $w = 3$ holds the latest frame of each window a two-window task is moving between, and drops the frames that would only compete with them. The limit is not set by the model: with up to 112 images in a single request, no request in the $w = \infty$ arm exceeded the model's limits. Because both arms also turn off the state signals, part of the loss comes from them, and this ablation does not separate the two. The number of tasks stopped at the 100-action limit points the same way. Of the 93 tasks, 2 reach the limit with the full agent, 3 without the state signals at $w = \infty$, and 7 without the state signals at $w = 1$. Without the budget notice the model no longer knows how much of its budget is left, and with $w = 1$ it also has to look again at windows it has lost sight of, which spends actions on finding its place rather than on the task.

\paragraph{Verifying After Acting.} Without the three checks the agent loses 2 points, one in each domain, but neither lost task traces to a missing check. In the Writer task, the request was to convert uppercase text to lowercase. The agent converted only the passages written entirely in capitals and confirmed the change in the saved file, while the reference document lowercases every letter. In the GIMP task, the agent flipped the image, saved it and checked the saved file. The score was lost afterwards, when the checker's own scripted export from the application produced no output file. The same GIMP task also failed in an earlier run of the full agent. With one run per configuration, a difference of two tasks is within run-to-run variation.

\begin{table}[t]
  \centering
  \footnotesize
  \setlength{\tabcolsep}{3pt}
  \caption{How the agent works with and without Verifying After Acting (one run per configuration). GUI actions count toward the 100-action budget. Looks are separate requests to see the screen, and model calls are the executor's. Input tokens, the share of them read from the prompt cache, and cost include the gate. Cost is estimated by the agent framework from the token counts at list prices. Capped is the number of runs stopped at the 100-action limit. The Writer full-agent run comes from an earlier version of the harness (Section~\ref{sec:ablation}).}
  \label{tab:process}
  \begin{tabular}{llrrrrrrrrr}
    \toprule
    Subset & Configuration & Score & GUI & Looks & Calls & \shortstack[r]{Input\\tok.\ (M)} & Cached & \shortstack[r]{Cost\\(\$)} & \shortstack[r]{Time per\\task (s)} & Capped \tabularnewline
    \midrule
    \multirow{2}{*}{Writer (23)}
      & Full & 23.00 & 431 & 37 & 555 & 26.5 & 79\% & 52.93 & 335 & 1 \tabularnewline
      & w/o Verifying & 22.00 & 295 & 196 & 589 & 25.6 & 72\% & 61.54 & 316 & 0 \tabularnewline
    \midrule
    \multirow{2}{*}{GIMP (26)}
      & Full & 26.00 & 201 & 61 & 331 & 13.9 & 75\% & 30.85 & 210 & 0 \tabularnewline
      & w/o Verifying & 25.00 & 226 & 195 & 513 & 20.6 & 72\% & 48.40 & 206 & 0 \tabularnewline
    \bottomrule
  \end{tabular}
\end{table}

What the checks change is how often the agent has to look (Table~\ref{tab:process}). For both domains we compare the ablated run with a full-agent run of the same model and agent framework. Without the capture, the agent asked to see the screen 195 times instead of 61 on GIMP and 196 times instead of 37 on Writer, or 0.86 and 0.66 times per GUI action instead of 0.30 and 0.09. Nine GIMP tasks and one Writer task were declared infeasible by the gate and end before the executor starts in both runs, and on every one of the other 17 GIMP and 22 Writer tasks the agent looked more often without the capture. Looking does not count toward the GUI action budget, so the extra looks cost model calls and tokens rather than budget. On GIMP, GUI actions changed little (226 against 201), while model calls rose from 331 to 513, input tokens by 48\% and estimated cost by 57\%, at about the same wall-clock time per task (206\,s against 210\,s). On Writer the totals compare less directly. In three tasks the full agent took 58, 30 and 100 GUI actions, mostly single key presses, against 10, 4 and 75 for the ablated agent, so the full run has more GUI actions in total (431 against 295) and one run that stopped at the 100-action limit, although that run still passed. On the other 19 feasible Writer tasks, model calls rose from 349 to 426, input tokens by 15\%, estimated cost by 21\% and time per task from 221\,s to 269\,s. The Writer full-agent run also comes from an earlier version of the harness, which did not yet show the step budget notice and passed skill rules to the executor in a different form, so it is a less controlled comparison than GIMP's. This is the separate turn that Section~\ref{sec:verify} describes: without the capture the model still verifies its actions, but pays a turn for each check. In both domains the ablation therefore shows Verifying After Acting making verification cheaper. One run per configuration is too few to show whether it also changes the score.

\needspace{8\baselineskip}
\section{Limitations}\label{sec:limitations}
Our results come from a single model, Claude Opus~5. The model-matched comparison in Section~\ref{sec:results} isolates the harness for this model only, and we have not measured how much the harness adds for weaker or different models. They also come from single runs. Some tasks pass and fail across repeated runs of the same configuration, so differences of one or two tasks are within run-to-run variation. The ablations are narrower still. Each runs on its own subset of domains, so they cannot be compared with one another, and none of them covers the full task set. The State and Context Management ablation also turns off all state signals together and changes the screenshot window at the same time, so it does not separate the contribution of the individual signals from that of the window.

The mechanisms themselves rest on what the harness can observe, which is less than the model does. The progress state is inferred from each call and its reply. A save is recognised only by its keyboard shortcut, so a save made through a menu is not seen and can be followed by a reminder the model does not need. Repeated failures are recognised only when a call and its reply repeat exactly, so a model that varies its arguments slightly while making the same mistake is not caught. The gate is held to its role partly by instruction. Its probe shell is read-only only because it is told to be, and nothing in the harness stops it from writing. It ends a task only when its verdict comes with findings and a reason, but the harness checks that these are present, not that they are correct, so a wrong verdict stated with confidence can still end a feasible task. Finally, we measure tokens, model calls and time per task only for the Verifying After Acting ablation, so the cost of the context window and of the gate is not quantified.

\section{Conclusion}\label{sec:conclusion}

We proposed a set of harness mechanisms, implemented on openJiuwen's agent-core, that address three challenges of long-horizon computer use: state and context drift, the knowledge gap, and error accumulation. For state and context drift, State and Context Management keeps task-wide facts in fixed positions, bounds the screenshots in view, and tracks for the model the budget, unsaved work and repeated failures it cannot see. For the knowledge gap, Probing Before Acting checks the task against the machine in a read-only pass before the first action, decides feasibility only with evidence, and passes the conventions the task depends on to the executor. For error accumulation, Verifying After Acting aims every input in the pixels the model sees, delivers it only to the window it names, and returns an explicit mark with a fresh capture, so an error surfaces at the step where it happens. All three sit in the harness around a fixed model, with no training and no extra model calls in the main loop. Evaluation on OSWorld-Verified shows a success rate of 93.5\%, 10.1 percentage points above Anthropic's own agent for the same model and 3.3 points above the strongest leaderboard entry, and removing any one mechanism deteriorates the system's performance, in accuracy or in tokens. Together, these results show that for a capable model, the harness that decides what the model sees, what it knows before it acts and what it learns after each action is a substantial lever on long-horizon computer use.

All of our code is open-sourced at: \url{https://github.com/openJiuwen-ai/computer-use/tree/osworld-challenge}

\clearpage
\bibliography{references}
\bibliographystyle{plainnat}

\clearpage
\appendix

\section{Author List}
\label{app:authors}

\textbf{Core Contributors.} Tao Yu, Xinyu Zhang, Kwangyang Chia, Zhiwen Huang, Xiaoneng Xiang, Yangkai Ding, Ruifeng Shi, Zheng Wang, Gan Bingzheng, Lan Hu, Tuo Xu, Hongbo Wang, Qi Ye, Xuefeng Jin, Zhangchun Zhao 
\end{document}